# On Optimizing Deep Convolutional Neural Networks by Evolutionary Computing

M. U. B. Dias[1], D. D. N. De Silva[2], S. Fernando[3]

[1,2,3]Department of Computational Mathematics, University of Moratuwa, Sri Lanka

*Abstract*

*Optimization for deep networks is currently a very active area of research. As neural networks become deeper, the ability in manually optimizing the network becomes harder. Mini-batch normalization, identification of effective respective fields, momentum updates, introduction of residual blocks, learning rate adoption, etc. have been proposed to speed up the rate of convergent in manual training process while keeping the higher accuracy level. However, the problem of finding optimal topological structure for a given problem is becoming a challenging task need to be addressed immediately. Few researchers have attempted to optimize the network structure using evolutionary computing approaches. Among them, few have successfully evolved networks with reinforcement learning and long-short-term memory. A very few has applied evolutionary programming into deep convolution neural networks. These attempts are mainly evolved the network structure and then subsequently optimized the hyper-parameters of the network. However, a mechanism to evolve the deep network structure under the techniques currently being practiced in manual process is still absent. Incorporation of such techniques into chromosomes level of evolutionary computing, certainly can take us to better topological deep structures. The paper concludes by identifying the gap between evolutionary based deep neural networks and deep neural networks. Further, it proposes some insights for optimizing deep neural networks using evolutionary computing techniques.*

## 1. Introduction

Training an Artificial Neural Network (ANN) for a given task is a still a demanding research topic in the field of Artificial Intelligence. In order to obtain the highest performance of the network, multi-dimensional optimization is essential which increase the complexity of the problem. The performance of ANN is determined by an aggregation of learning rate, accuracy and generalization. Learning rate is very important as providing an enormous labeled dataset for training. On other hand, the time taken to achieve a given benchmark of accuracy should be reduced as much as possible. Therefore, the learning rate should be maintained properly in order to enhance the performance of ANN. The overall learning rate can be measured by number of epochs and time taken for learning. The accuracy of the ANN is another key component of the performance since the output should be with minimum error after a certain number of epochs. If the final error is beyond the desired margin of error, then the ANN is with poor performance. The error of an ANN is measured by mean squared error, mean absolute error, exponential error method, etc. After all, the designed ANN should be able to generalize well. Generalization means the characteristic of applicability to any problem within a given scenario. In a nutshell, how well it will be able to give the desired output when a new input is given. This generalization is related with the problem of over fitting of ANN. If the ANN is trained more for a particular type of uniform dataset, the accuracy will be substantially increasing for that particular type of input data points. But, when future data point is given, the output will be worst. This incident is called as, over fitting of a network. Therefore, generalization of ANN should be protected throughout the designing and training process. Ultimately, ANN performance can be defined as follows; any well-generalized ANN should be attained for a given subset of problems with high accuracy in less number of epochs.

In order to maximize the performance of ANN, the determination of the optimal network structure and weights for the given task, is a multi-dimensional problem with a vast space of solutions. The process of determining the most appropriate topology of network for the given task is called as structural optimization. The number of hidden neurons and the connectivity between these neurons construct the topology of the network. Some neurons have to be dropped because of their less effectiveness for the network and some neurons have to be added due to their usefulness. Then the feed forward and recurrent connectivity between these neurons have to be determined in order to construct the complete structure. This connectivity is not about the weight of the connection, but the existence of the particular connection. The structure is usually decided using an ad-hoc choice or some trial and error method. The ad-hoc choice is done by a priori analysis of the task. This is much complicated and non-deterministic due to the lack of a proper model of the task. Furthermore, a conventional ad-hoc structure does not provide the optimum solution for a given task. Apparently, shoehorning the given task to a pre-defined structure is inappropriate; rather the structure should be shoehorned to the given task. The trial and error methods can be categorized as constructive and destructive methods. Constructive methods are initiated with a simple and small structure and later more neurons and connections are added to improve the performance. In contrast, destructive methods initiated with complex and massive network and gradually delete the connections and neurons to make the structure simple and small. These time and error methods are computationally prohibitive and more likely to be trapped in structural local optima because of the stochastic search methods used. Moreover, these search methods are limited for some predefined topologies; hence search through a narrow solution space only. The evolutionary computing comes on stage in this context. Evolutionary computing methods have become successful alternative for topological optimization, due to larger search space, higher speed and more probability to achieve global optima. By using the genetic algorithms and evolutionary programming, some algorithms have outperformed the conventional structural learning methods.

The process of determining the weights of the connections is called as parametric optimization. The weights represent the strength of the connections. These weights are usually found by Stochastic Gradient Descent (SGD) algorithm. The convergent rate and the likelihood to trap in local optima are drawbacks of this method. As per the literature, few researches have been done to apply evolutionary computing methods in parametric optimization.

## 2. Evolutionary Computing on ANN Training

The research into applying Evolutionary Computing to improve the performance of ANN can be mainly explained under the strategies as representing network in terms of individuals of the population, initialization of population by initial network structure, adoption of proper fitness functions to find optimum network structure, development of parent selection criteria to produce better structure, and reproduction operators. The next couple of sections describe those strategies in detail.

*A. Representation of Individuals*

In [1], an individual chromosome is interpreted with a bit string which is a combination of several sub strings. The number of bits of a sub string is decided by the granularity. A coarse granularity has a very narrow search space, hence less computational cost. A fine granularity has a large search space, but computationally expensive. Therefore, the number of bits in a sub string is a critical parameter that should be decided corresponding to the given task and network. The first substring of the chromosome indicates the granularity of the string, in binary numbers. Then all other substrings represent the connection links between the neurons. First bit of a substring represents the connectivity. If the connection exists it has value '1' and if the connection does not exist it has value '0'. The rest of the bits of a substring interpret the weight using a binary encoding. Particularly, if the granularity is n, the different number of weight values that can be represented is $2^{n-1}$. For example, if the number of bits is 4, it can represent 8 different weight values, such as -2 to 5 with binary encoding 000 ~111 (i.e. -2=000, -1=001, 0=010... 5=111). These weight values are included only when connection are existing. That means, if the connectivity bit is 0 the rest of the sub string will be disappeared. Therefore, the different individuals may have different length of strings. Finally, these substrings are arranged in an order, such a way to keep the substrings, which represent the connections coming to the same neuron, at nearby.

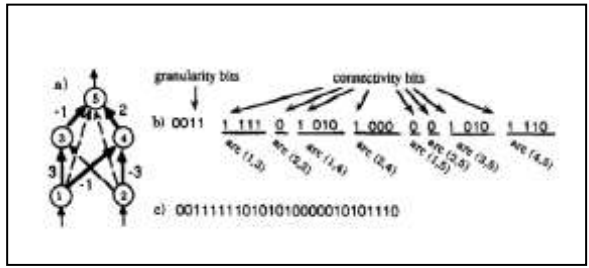

Figure 1: Representation of chromosome in [1]

There are some drawbacks with this representation. As mentioned above the individuals may have different bit lengths and so that difficult to mate two of them straight away. Besides that, this is limited for some range of weight values; consequently, a small search space is utilized. Furthermore, this encoding scheme is not applicable to recurrent networks. Moreover, this representation does not allow add/delete neurons, hence the number of neurons is constant throughout the process. However, the other researches have demonstrated the importance of changing this parameter (i.e. number of hidden neurons). [19] is also use a similar type of genotype, which consist two types of genes. Neuron genes represent the input, output and hidden neurons while connection genes represent the connectivity. Each and every neuron has a particular number that want change throughout the process. The connection genes carry the information of connection input and output neuron number, weight, the availability of connection (connectivity bit), and innovation number (this is explained later).

Since evolutionary programming usually does not use an encoding scheme, [2] and [3] have used real number values to represent the individuals. If the number of input and output neurons are m and n respectively, and maximum number of hidden neurons is N, then size of (m+N+n) x (m+N+n) two matrices are used. One is a binary matrix, which represents the connectivity between the neurons, and other one is weight matrix, which represents the values of particular connections. In [3] these two matrices are upper triangular matrices, because the presented algorithm doesn't applicable for recurrent neural networks. Then, another N dimension vector is used to denote the existence of the hidden neurons. The components of the vector can be either value 1 or 0. If a particular neuron exists it indicates as value '1' and if it doesn't exist it indicates as '0'. Since evolutionary programming uses asexual reproduction, only (no crossover operations) this notation suits well.

*B. Initialization of the Population*

Maniezzo [1] generates an initial population randomly, constrained to the given range of granularity and the given number of neurons. The other researchers [2] and [3] randomly select the number of hidden neurons and the number of connection links, from uniform distributions over user defined ranges. Then, the weight values are also generated from a uniform distribution over a small range. [3] does a further modification after generating the initial population. It trains the population partially using BP and then if the error is reduced in a particular individual then it is marked as "success" or otherwise as "failure". [19] initialize the networks without hidden neurons. The population begins with the simplest network and the hidden neuron are added according to the performance.

*C. Fitness Function*

For supervised learning there are three possible ways to measure the fitness [2]. Summation of square error (1), Summation of absolute error (2), or Summation of exponential error (3) of the n[th] individual can be used to measure the fitness of the particular individual. $t_i$ is the targeted output of i[th] labeled data point used for training, and $a_i$ is the actual output for the particular input. The superscript $\lambda$ denotes an individual of the population.

$$E^\lambda_{sqr} = \sum_i (t_i^\lambda - a_i^\lambda)^2 \qquad (1)$$

$$E^\lambda_{abs} = \sum_i |t_i^\lambda - a_i^\lambda| \qquad (2)$$

$$E^\lambda_{exp} = \sum_i e^{|t_i^\lambda - a_i^\lambda|} \qquad (3)$$

Yao and et al [3] uses a more advance method to evaluate the individuals. It calculates the mean square error percentage and normalizes by the factors of number of output neurons and the range of the output values. This method is adopted from [4]. If the number of output neurons is *n* the number of data points used for validation is T, and the maximum and minimum values of the outputs are $O_{max}$ and $O_{min}$, the error for individual $\lambda$ can be calculated as follows (4);

$$E^\lambda = 100 \cdot \frac{O_{max}^\lambda - O_{min}^\lambda}{n} \cdot \frac{E_{sqr}^\lambda}{T} \quad (4)$$

*D. Parent Selection*

Maniezzo [1] has used a very primary parent selection method; i.e. randomly pair the individuals of the existing population. Then, each and every pair will undergo a crossover operation. The mutation operations are implemented on each individual with a pre-defined probability. However, [3] uses a rank based method to select a parent. Every individual of the population is given a rank from 0 to (M-1) from the descending order of the fitness, i.e. the ascending order of the calculated error. It has reduced the computational load by avoiding taking the reciprocal of the error to calculate the fitness value. The probability of the $\lambda$ th individual to be selected is (5);

$$P(\lambda) = \frac{M - \lambda}{\sum_{k=1}^{M} k} \quad (5)$$

The main drawback of this method is; it assumes a uniform distribution of error values, even though it is not. The others [2] use a very simple method to select the parents. The fittest half of the existing population is used to produce the offspring of the next generation.

*E. Reproduction Operators*

Each and every parent undergoes for 3 type of mutations in [1]. Granularity bit mutations, connectivity bit mutations, and weight bit mutations are executed with three user defined mutation probabilities. The n point crossover operator is used for sexual reproduction. Although individuals may have different lengths of bit strings to represent them, all of them are stored in maximum possible fixed length memories. That is the string length of an individual with maximum possible granularity and maximum connectivity. For example, if the maximum possible connections are 5 and maximum granularity is 3 then, then the maximum possible length will be 22, with two bits to represent the granularity. The individuals with less number of bits just don't use the rest of bits to represent their network. The crossover operator is implemented using these fixed length bit strings, so that no problem of mating different size of individuals. However, one drawback of this operator is the lacking of compositionality property. Compositionality is the meaningfulness of a portion of a string.

Some researchers [2], [3] use asexual reproduction only. Two types of mutations are performed in order to obtain a valued offspring; i.e. parametric mutations and structural mutations. Rather using fixed probabilities like [1], [2] yield the probabilities by uniform distributions. First it calculates a 'temperature value' (T) for particular parent using its fitness value. If the maximum attainable fitness value is $f_{max}$ and fitness of the $\lambda$ individual is $f_\lambda$ then,

$$T_\lambda = 1 - \frac{f_\lambda}{f_{max}} \quad (6)$$

gives the temperature value for that parent. Then it calculates an instantaneous temperature value ($\widehat{T}_\lambda$) for every mutation operator implementation.

$$\widehat{T}_\lambda = U(0,1) T_\lambda \quad (7)$$

where $U(0,1)$ is a random variable chosen from a uniform distribution over the interval [0,1]. The weights are updated choosing a random variable from a normal distribution;

$$W_{new} = W_{old} + N(0, \alpha \widehat{T}_\lambda) \quad (8)$$

where, $\alpha$ is a user-defined proportion. There are four kinds of structural mutations used in [2] and [3], those are; adding a hidden neuron, deleting a hidden neuron, adding a connection link and deleting a connection link. [2] apply these mutations on parents with particular number calculated using pre-defined interval of $[\Delta_{min}, \Delta_{max}]$ for each four structural mutation type. This particular number for each individual is given in eq.(9).

$$\Delta_{min} + [U(0,1)\widehat{T}_\lambda(\Delta_{max} - \Delta_{min})] \quad (9)$$

The researcher [3] uses a hybrid training method of back propagation(BP) and simulated annealing(SA), for parametric mutation. When a parent is selected, if it is marked as "success" then it undergoes for further BP training and no further any mutation is done. Else if it is marked as "failure" then it is trained using SA and update whether success or failure. If it's success then no further mutations are performed, but if it's still "failure" structural mutations are performed followed by partial BP training in the sequence of hidden neurons deletion, connection links deletion, connection links addition and hidden neuron addition. These structural mutations are executed according to the survivor selection method described in the next subtopic. The number of hidden neurons and connection links to be deleted is random values chosen from small uniform distributions defined by the practitioner. These generated numbers of neurons are deleted uniformly at random. The connection links are deleted according to the importance of a connection. The importance of a connection is evaluated by a variable called 'test' of that particular connection. When T number of validation patterns is used, if the particular connection has a weight of $'w'$ then;

$$test(w) = \frac{\sum_{i=1}^{T} \xi^i}{\sqrt{\sum_{i=1}^{T}(\xi^i - \overline{\xi})^2}} \quad (10)$$

where $\xi^i = w + \Delta w^i$ and $\Delta w^i = -\eta \frac{\partial E_{abs}}{\partial w}$.
$\overline{\xi}$ is the average value of $\xi^i$ over T number of validation patterns. According to this test value, the connections are deleted. Same value is used for addition of connection links and these adding connections are selected from the connections, which are currently with zero weights. In neuron addition, a process called "cell division" is used. An existing neuron is spitted into two parts and following weight vector updating is carried out.

$$\begin{aligned} w_{ij}^1 &= w_{ij}^2 = w_{ij} & i \geq j \\ w_{ki}^1 &= (1 + \alpha) w_{ki} & k > i \\ w_{ki}^2 &= -\alpha w_{ki} & k > i \end{aligned} \quad (10)$$

Where $w^1$ and $w^2$ are weight vectors of new neurons and $\alpha$ is a mutation parameter which may be either fixed value or random value.

One of the main disadvantages of these structural mutations is the generational gap. That means the huge behavioural differences between the parents and the offspring. [2] introduces the new hidden neurons without any connections and connections with zero weights preventing radical jumps in fitness values in these two types of mutations. But, it claims that in addition of connection links/neurons these sudden changes are inevitable. After all four types of structural mutations, [3] use a partial training with BP to avoid these sudden behavioural changes. Additionally, the added connection links are initialized with small weight values in [3], in contrast to zero initial weights in [2], because of the partial training with BP.

[19] uses two structural mutations only; i.e. addition of connection and addition of hidden neuron. Connections are added with random weight values. The new hidden neurons are added by splitting the existing connections. The new neuron will get an input connection with a weight of 1 and output connection with weight of old connection weight. Preserving the old weight value will reduce the generational gap between the parents and the offspring. Whenever one of these mutations are occurred, a new gene is added to the chromosome, which leads to vary the size among the individuals. Every new gene is given a number called "innovation number" which is incremented in every single mutation. This number of a particular gene, want change in the entire process. Hence, the historical data will be preserved and can be utilized whenever needed. This feature is not available in any other method. These data is used to line up the individuals with different sizes in order to implement the crossover operation.

## 3. Optimization Techniques in Deep Convolution Networks

The techniques that have been used in shallow neural networks have been further advanced and applied in deep neural networks. Especially, normalization techniques which were used to normalize the inputs in conventional neural networks, has been extended to weight normalization as well as intermediate output normalization. Further, different momentum algorithms have been derived to speed up the convergence of deep networks, the next couple of sections sum up the common approaches that have been practiced to smooth and speed up the learning process of deep networks. Here, the paper focuses only on convolutional neural networks (CNN) because it has been recognized as a key approach for object recognition.

*A. Normalization Techniques*

In deep networks, input to each layer is affected by parameters of previous layers, as network parameters changes (by training), even small changes to the network get amplified down the network. This leads to change in the statistical distribution of inputs to following layers from previous layers, therefore hidden layers will keep trying to adapt to that new distribution, hence slows down the convergence and make it difficult for training. This may lead to requirement of low learning rates and careful parameter initialization. This is known as internal covariant shift. As a general solution, it introduced to normalize the data (by mean and variance) and several normalization techniques have been tried out.

*1) Normalization*

Normalize data by both mean and variance is a major technique, which simply is to transform to make data with zero mean and unit variance hence de-correlated, through a series of linear transformations. The process centers, the data around value zero by subtracting the mean and then divide by the standard deviation for scaling. In general, subtracting the mean across every individual feature in the data, and make geometric interpretation of centering the cloud of data, around the origin along every dimension (k). To normalize the data dimensions so that they are of approximately the same scale, divide each dimension (k) by its standard deviation after they have been zero centered. This is also known as simplified whitening process. This simplified whitening only removes mostly the first order covariant shift, but for removing higher order shift requires complex techniques have been introduced.

*2) Local contrastive normalization*

Local contrastive normalization (LCN) performs a local competition among adjacent features (like pixels) in feature maps and between features at the same spatial location on different feature maps. LCN applies after introducing the non-linearity (ReLu) for whitened data.

For example, let us consider a local field of 3x3-area portion (9 pixels) to clarify the process. First, for each pixel in a feature map, find its adjacent pixels (radius is 1 in this case), so there are 8 pixels around the target pixel in the middle (do the zero padding if the target is at the edge of the feature map). Then, compute the mean of these 9 pixels (8 neighbor pixels and the target pixel itself), subtract the mean for each one of the 9 pixels. Next, compute the standard deviation of these 9 pixels. In addition, judge whether the standard deviation is larger than 1. If larger than 1, divide the target pixel's value by the standard deviation. Otherwise, keep the target value as what they are (after mean subtraction). At last, save the target pixel value to the same spatial position of a blank feature map as the input of the next layer of the deep CNN.

LCN introduces a competition among the output of adjacent convolution kernel. This normalization technique is useful when it deals with ReLU neurons because ReLU neurons have unbounded activations and needs local responsive normalization to normalize that. For detecting high frequency features with a large response, normalizing around the local neighborhood of the excited neuron, it becomes even more sensitive as compared to its neighbors [12].

*3) Batch Normalization*

Batch normalization (BN) is a learnable whitening process that normalizes the inputs to each following hidden layer so that their distribution kept fairly constant as training proceeds, hence improves training and allows faster convergence. About input distributions, BN algorithm addresses the changing distributions issue which known as internal covariant shift and allows using higher learning rates. These learnable hyper parameters in BN transformation are learned through back propagation during online or mini-batch training. Furthermore, Batch normalization reduces effects of exploding and vanishing gradients while regularize the model. Without BN, low activations of one layer can lead to lower activations in the next layer, and then even lower ones in the next layer and so on [11].

At the beginning batch normalization initialize (with mini batches) by normalizing the data using calculated mini batch mean and variance hence standard deviation Not just normalizing each input of a layer may change what the layer can represent. To address this, it introduced a transformation, for each normalization, a pair of parameters, which scale and shift the normalized value. These parameters are learnable during the training using back propagation and by setting them equal to standard deviation and mean respectively it could even recover original activations. These learned parameters in transformation depend on all the training examples in the mini batch. Also, learning ensures that as the model trains, layers can continue learning on input distributions that exhibit less internal covariate shift, thus accelerating the training.

B. *Momentum Updates*

   1) *Vanilla momentum update*

Momentum is one of the key approaches that have been applied to faster the convergence of deep networks getting out of local minima. With momentum m, the weight update at a given time t, m adds a fraction of the previous weight update to the current one as shown in eq.20. When the gradient keeps moving into the same direction, m increases the size of the steps toward the minima. On contrast, when the gradient changes the direction compared to previous few steps, momentum help to smooth out variation.

$$\Delta w(t) = -\alpha . \frac{\partial E}{\partial w} + m.\Delta w(t-1) \quad (20)$$

   2) *Nesterov momentum update*

Nesterov momentum is a derivative of momentum updates, which performs well in convex functions as in eq.21 and eq.22. The basic idea of Nesterov momentum is to compute the gradient of the future approximate position than the current position of the parameter. This accelerated momentum helps to rush the network to its convergence [5, 9].

$$w^* = w(t) + m.\Delta w(t-1) \quad (21)$$

$$\Delta w(t) = -\alpha . \frac{\partial E}{\partial w^*} + m.\Delta w(t-1) \quad (22)$$

C. *Effective Receptive Field*

Particularly in CNN, behaviour of a unit depends only on a region of the input, which is called a receptive filed of the unit. Theoretically, size of the receptive field of an output unit can be increased by stacking more layers to make much deeper network, or by sub-sampling, which increases the size of receptive field. Deeper analysis into receptive fields has shown that pixels in a receptive field do not contribute equally to response of the output unit. Indeed, the central pixels in the receptive field make much larger impact on the response rather than the boundary pixels. These pixels, which make larger impact on the output, is known as effective receptive field (ERF) [6]. Further, these effective receptive fields are Gaussian distributed and growing bigger in size during the training. Since ERF are smaller at initial stages, and causes problems for the tasks that require a larger receptive field, the weights are suggested to initialize in lower scale at the center and in higher scale at the outside. However, an architectural change to CNN has also been suggested in terms of defining convolution window which emphasis the need of defining window, which connect larger area with some sparse connection, rather than defining small local convolution window [10].

*http://www.erogol.com/dilated-convolution/*

D. *Dropout training*

Model combination always improves the performance of neural networks. With large neural networks, averaging the outputs of many separately trained nets is expensive. Training many different architectures is hard, because finding optimal hyper-parameters requires a lot of computation. Large networks normally require large amounts of training data and there may not be enough data available to train different networks on different subsets of the data. Dropout is a technique that prevents over-fitting and provides a way of approximately combining exponentially many different neural network architectures efficiently. Dropout refers to dropping out units (hidden and visible) in a neural network. Training a network with dropout and using the approximate averaging method at test time has led to significantly lower generalization error [15].

E. *Annealing Learning Rate*

In deep networks, many researchers have tried to anneal the learning rate over time. Knowing when to decay the learning rate is very tricky. Decaying the learning rate slowly can cost high computational time and on the other hand decaying the learning rate aggressively may not help the system to reach the global minimum. Step decay, and exponential decay are common types of implementing the decay. Step decay reduces the learning rate by some factor every few epochs. Exponential decay reduces the learning rate exponentially over the number of iteration. Adaptive learning rates such as Adagrad [13], RMSprop and Adam [14] are some of the significant learning rates that have been proposed to improve the performance of Deep neural networks.

## 4. Evolutionary Strategies in Deep Convolution Networks

Optimizing topological structure of deep neural networks has become more challenging task as the increase of the depth of the network. No longer it is feasible to find the optimum structure for deep networks for a given problem manually and has opened up new branch in deep networks to look into the possibility of automating the task. As we discuss in early sections, many approaches have been proposed to optimize the deep networks, however Evolutionary Computing (EC) based approaches are being called back to overcome the difficulty. Application of EC into reinforcement learning on Atari games [17], and EC into Long-Short-Term-Memory (LSTM) on image capturing and language modeling [16] are some of the significant recent applications which could confirm the possibility of improving efficiency of deep networks using EC based techniques.

Among them, few researchers have attempted to improve the topological structure of deep CNN. Miikkulainen et al [16] has evolved a CNN network which could reach the convergence after 30 epochs and it is compared to 4 times faster than the manually defined approach for the same problem set (CIFAR-10). In their approach, a convolution layer has been defined as a node in chromosome in evolutionary programming. Problem was solved as multi-population problem that evolves structure and then hyper-parameters for the assembled network. However, in this approach optimizing of hyper-parameters were done after the network structure was assembled by genetic programming. Suganuma et al[18] has applied Cartesian genetic programming to evolve the CNN in image classification task. In addition to convolution blocks, batch normalization and residual blocks have been considered in evolving the topological structure. However, the hyper-parameters of the network have been optimized after assembling the network.

## 5. Discussion

As the performances of deep neural networks increases, in general the number of layers in deep neural networks is also increases. That is the more the number of layers; the more accuracy has been recorded. Optimizing the deep neural network's topological structure for a given data set has become a challenging task for neural network researchers. Therefore, many researchers are now exploring the possibility of applying genetic programming or evolutionary algorithm to evolve deep neural networks.

In conventional neural networks, this has been mainly achieved by applying evolutionary algorithm. In there, the network hyper-parameters are encoded into genes of chromosomes, and then have tried to optimize the parameters based on a defined fitness-function. Or else, the network topological structure has been tried to encode into chromosomes of genetic algorithm and tried to optimize the network topological structure. However, these approaches are not good enough to optimize the deep neural networks, because compared to conventional neural networks, deep neural networks has variety of architectures combing many different modules or layers. For example, residual layer, normalization layer, inception module, etc. Therefore, new approaches have been investigated to optimize the deep neural networks' topological structure or hyper-parameters. Few researches have encoded the modules of deep neural networks into chromosomes and tried optimize the neural network structure. However, those techniques are required further researches to evaluate their performances thoroughly.

Here the paper proposes to optimize the deep neural network structures as a collective intelligence of genetic programming: representing deep neural networks in terms of acyclic graph in detail (into neuronal level and connection) using genetic programming. As per the concept governed by swarm intelligence, each module or layer in the network is identified as a local genetic program. Hence the entire network can be considered as multiple genetic programs. Each entity is trying to optimize its behaviour locally under the constraint of optimizing overall behaviour of entire network. In this way, we can approach the deep neural network optimization problem, as emergent feature of collective multi-genetic programs.